\documentclass[11pt,a4paper]{article}
\usepackage[hyperref]{emnlp2018}
\usepackage{times}
\usepackage{latexsym}
\usepackage{url}
\usepackage{microtype}
\usepackage{subfiles}
\usepackage{graphicx}
\usepackage{float}
\usepackage{amsmath}
\usepackage{mathrsfs}
\usepackage{amsfonts}
\usepackage{amssymb}
\usepackage{multirow,tabularx}
\usepackage{color}
\usepackage{inconsolata}
\usepackage{listings}
\usepackage{booktabs}
\usepackage[skip=5pt]{caption}
\usepackage{makecell}
\usepackage[linesnumbered]{algorithm2e}
\usepackage[font=small]{caption}
\usepackage{subcaption}

\usepackage{color}

\definecolor{mygreen}{rgb}{0,0.6,0}
\definecolor{mygray}{rgb}{0.5,0.5,0.5}
\definecolor{mymauve}{rgb}{0.58,0,0.82}

\definecolor{javaGreen}{RGB}{63,127,95}

\aclfinalcopy 

\title{Mapping Language to Code in Programmatic Context}

\author{
Srinivasan Iyer$^*$, Ioannis Konstas$\dagger$, Alvin Cheung$^*$ and Luke Zettlemoyer$^*$\\
$^*$Paul G. Allen School of Computer Science and Engineering, Univ. of Washington, Seattle, WA \\
\tt{\{sviyer, akcheung, lsz\}@cs.washington.edu} \\
$\dagger$Heriot-Watt University, Edinburgh, UK\\
\tt{i.konstas@hw.ac.uk}\\
}\date{}

\begin{document}
\maketitle
\begin{abstract}

Source code is rarely written in isolation. It depends significantly on the programmatic context, such as the class that the code would reside in. To study this phenomenon, we introduce the task of generating class member functions given English documentation and the programmatic context provided by the rest of the class. 
This task is challenging because the desired code can vary greatly depending on the functionality the class provides (e.g., a sort function may or may not be available when we are asked to ``return the smallest element'' in a particular member variable list).  
We introduce CONCODE, a new large dataset with over 100,000 examples consisting of Java classes from online code repositories, and develop a new encoder-decoder architecture that models the interaction between the method documentation and the class environment. 
We also present a detailed error analysis suggesting that there is significant room for future work on this task. 

\end{abstract}

\section{Introduction}

Natural language can be used to define complex computations that reuse the functionality of rich, existing code bases. However, existing approaches for automatically mapping natural language (NL) to executable code have considered limited language or code environments. They either assume fixed code templates 
(i.e., generate only parts of a method with a predefined structure;
\citeauthor{quirk-mooney-galley:2015:ACL-IJCNLP}, \citeyear{quirk-mooney-galley:2015:ACL-IJCNLP}), a fixed context (i.e., generate the body of the same method within a single fixed class; \citeauthor{ling2016}, \citeyear{ling2016}),  or no context at all (i.e., generate code tokens from the text alone; \citeauthor{oda2015learning}, \citeyear{oda2015learning}). 
In this paper, we introduce new data and methods for learning to map language to source code within the context of a real-world programming environment, with application to generating member functions from documentation for automatically collected Java class environments.

\begin{figure}
    \includegraphics[width=\linewidth]{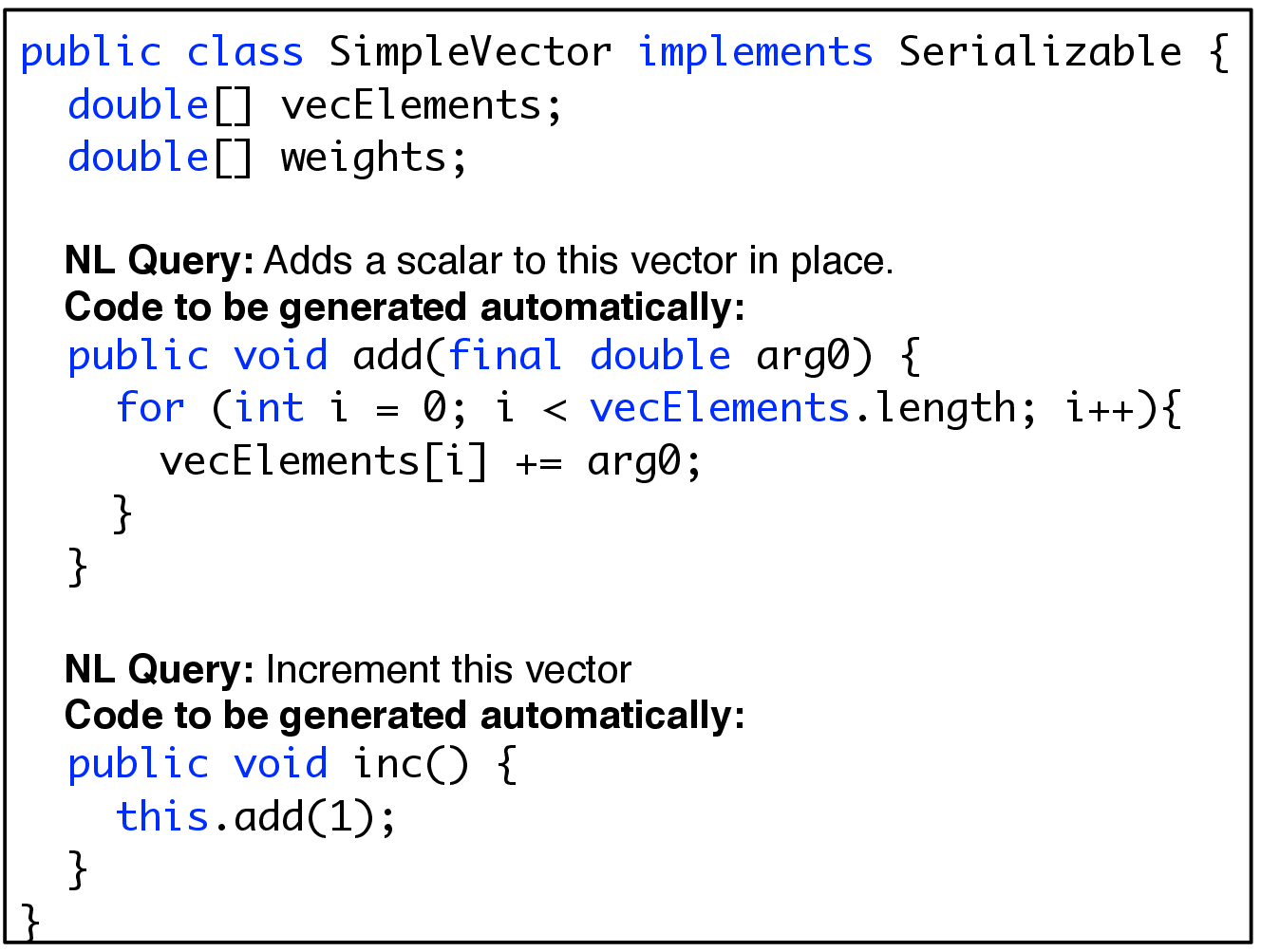}
    \caption{Code generation based on the class environment and method documentation. The figure shows a class where the programmer wants to automatically generate the \texttt{add} method from documentation, assuming the rest of the class is already written. The system needs to understand that {\tt vecElements} is the vector to be augmented, and that the method must take in a scalar parameter as the element to be added. The model also needs to disambiguate between the member variables {\tt vecElements} and {\tt weights}.}
    \label{fig:maineg}
\end{figure}

The presence of rich context provided by an existing code environment better approximates the way programmers capitalize on code re-use, and also introduces new language understanding challenges.
Models must (a) map the NL to environment variables, library API calls and user-defined methods found elsewhere in the class based on their names, types and signatures, and (b) decide on the structure of the resulting code.
For example, in Figure \ref{fig:maineg}, to generate the method \texttt{inc()} from the corresponding NL, \textit{Increment this vector}, it is crucial to know of the existence of class method \texttt{add()}. This helps us decide if it should directly call \texttt{add()} or generate the method from scratch by iterating through the \texttt{vecElements} array and incrementing each element. Similarly, for generating the \texttt{add()} method, the code needs to use the class variable \texttt{vecElements} correctly. Overall, the code environment provides rich information relating to the intent of the developer, and can be used to considerably reduce ambiguity in the NL documentation.

To learn such a code generator, we use a specialized neural encoder-decoder model that (a) encodes the NL together with representations based on sub-word units for environment identifiers (member variables, methods) and data types, and (b) decodes the resulting code using an attention mechanism with multiple steps, by first attending to the NL, and then to the variables and methods, thus also learning to copy variables and methods. This two-step attention helps the model to match words in the NL with representations of the identifiers in the environment. Rather than directly generating the output source code tokens \cite{dong2016,iyer-EtAl:2017:Long}, the decoder generates production rules from the grammar of the target programming language similar to \newcite{rabinovich-stern-klein:2017:Long}, \newcite{yin-neubig:2017:Long}, and \newcite{krishnamurthy-dasigi-gardner:2017:EMNLP2017} and therefore, guarantees the syntactic well-formedness of the output. 

To train our model, we collect and release CONCODE, a new dataset comprising over 100,000 (class environment, NL, code) tuples by gathering Java files containing method documentation from public Github repositories. This is an order of magnitude larger than existing datasets that map NL to source code for a general purpose language (MTG from \newcite{ling2016} has 13k examples), contains a larger variety of output code templates than existing datasets built for a specific domain, and is the first to condition on the environment of the output code. Also, by design, it contains examples from several domains, thus introducing open-domain challenges of new identifiers during test time (some e.g. class environments are LookupCommand, ColumnFileReader and ImageSequenceWriter). Our model achieves an exact match accuracy of 8.6\% and a BLEU score (a metric for partial credit; \citeauthor{papineni2002bleu}, \citeyear{papineni2002bleu}) of 22.11, outperforming retrieval and recent neural methods. We also provide an extensive ablative analysis, quantifying the contributions that come from the context and the model, and suggesting that our work opens up various areas for future investigation.


\section{Task Definition}

\begin{figure}[h]
    \includegraphics[width=\linewidth]{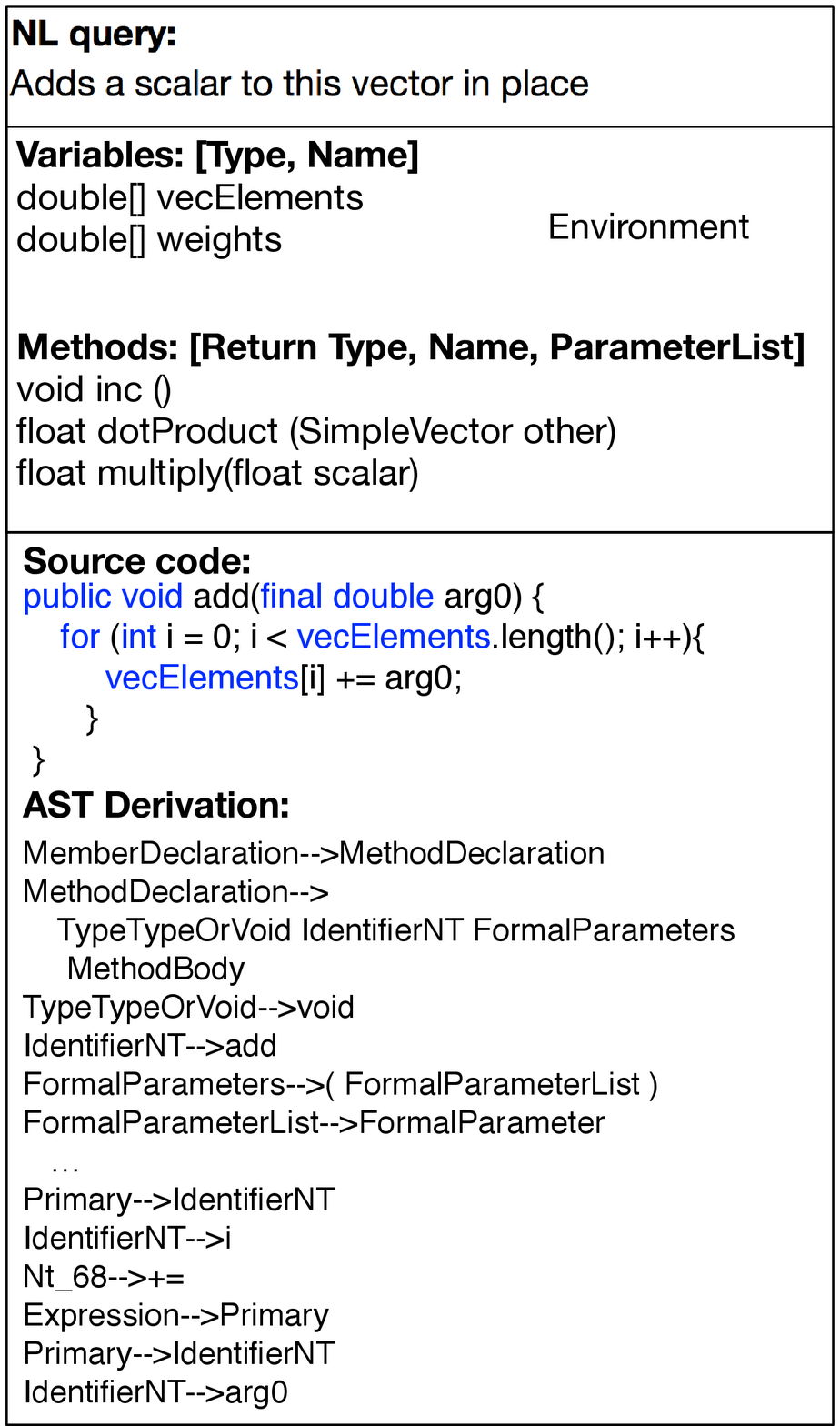}
    \caption{Our task involves generating the derivation of the source code of a method based on the NL documentation, class member variables (names and data types), and other class member methods (method names and return types), which form the code environment.}
    \label{fig:task}
\end{figure}

We introduce the task of generating source code from NL documentation, conditioned on the class environment the code resides in. The environment comprises two lists of entities: (1) class member variable names with their data types (for example, \texttt{double[] vecElements} as seen in Figure~\ref{fig:task}), and (2) member function names together with their return types (for example, \texttt{void inc()}).\footnote{The method parameters and body can be used as well but we leave this to future work.} Formally, let $q^{(i)}$, $a^{(i)}$ denote the NL and source code respectively for the $i^\text{th}$ training example, where 
$a^{(i)}$ is a sequence of production rules that forms the derivation of its underlying source code. The environment comprises a list of variables names  ${v^{(i)}}_{1..|v^{(i)}|}$ and their corresponding types  ${t^{(i)}}_{1..|t^{(i)}|}$, as well as method names ${m^{(i)}}_{1..|m^{(i)}|}$  and their return types ${r^{(i)}}_{1..|r^{(i)}|}$. Our goal is to generate the derivation of $a^{(i)}$ given $q^{(i)}$ and the environment (see Figure \ref{fig:task}). 

\section{Models}

We evaluate a number of encoder-decoder models that generate source code derivations from NL and the class environment. Our best model encodes all environment components broken down into sub-word units \cite{sennrich-haddow-birch:2016:P16-12} separately, using Bi-LSTMs and decodes these contextual representations to produce a sequence of valid production rules that derive syntactically valid source code. The decoder also uses a two-step attention mechanism to match words in the NL with environment components, and then uses a supervised copy mechanism \cite{gu-EtAl:2016:P16-1} to incorporate environment elements in the resulting code. We describe this architecture below.

\subsection{Encoder} 

\begin{figure}
    \includegraphics[width=\linewidth]{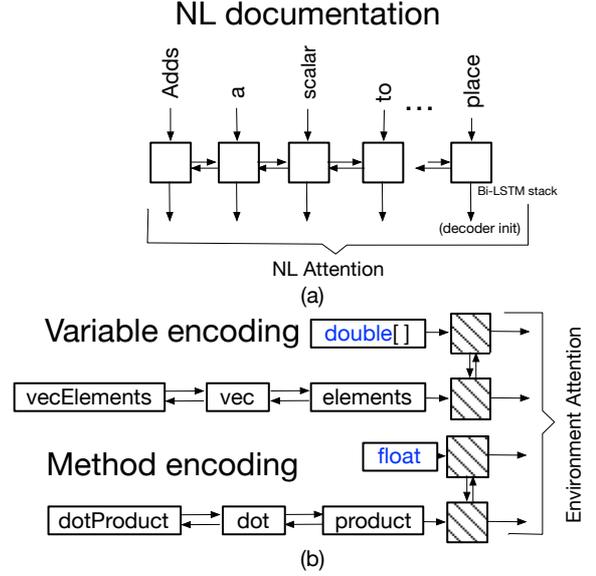}
    \caption{The encoder creates contextual representations of the NL (a), the variables and the methods (b). Variable (method) names are split based on camel-casing and encoded using a BiLSTM. The variable (method) type and name are further contextualized using another BiLSTM.}
    \label{fig:encoder}
\end{figure}

The encoder computes contextual representations of the NL and each component in the environment. Each word of the NL, $q_i$, is embedded into a high dimensional space using Identifier matrix $I$ (denoted as $\mathbf{q_i}$) followed by the application of a n-layer bidirectional LSTM \cite{hochreiter1997long}. The hidden states of the last layer ($h_{1}, \cdots, h_{z}$) are passed on to the attention layer, while the hidden states at the last token are used to initialize the decoder.
\begin{align*}
        h_{1}, \cdots, h_{z}  = \text{BiLSTM}(\mathbf{q_{1}}, \dots, \mathbf{q_{z}})
\end{align*}
To encode variables and methods, the variable types ($t_i$) and method return types ($r_i$) are embedded using a type matrix $T$ (denoted as $\mathbf{t_i}$ and $\mathbf{r_i}$). To encode the variable and method names ($v_i, m_i$), they are first split based on camel-casing, and each component is embedded using $I$, represented as $\mathbf{v_{i1}}, \dots, \mathbf{v_{ij}}$ and $\mathbf{m_{i1}}, \dots, \mathbf{m_{ik}}$. The encoded representation of the variable and method names is the final hidden state of the last layer of a Bi-LSTM over these embeddings ($\mathbf{v_i}$ and $\mathbf{m_i}$). Finally, a 2-step Bi-LSTM is executed on the concatenation of the variable type embedding and the variable name encoding. The corresponding hidden states form the final representations of the variable type and the variable name ($\hat{t_i}$, $\hat{v_i}$) and are passed on to the attention mechanism. Method return types and names are processed identically using the same Bi-LSTMs and embedding matrices ($\hat{r_i}$, $\hat{m_i}$).
\begin{align*}
        \mathbf{t_i} &= t_i T; \text{      } \mathbf{v_{ij}} = v_{ij} I   \\
        \mathbf{r_i} &= r_i T; \text{      } \mathbf{m_{ik}} = m_{ik} I \\
        \mathbf{v_i} &= \text{BiLSTM}(\mathbf{v_{i1}}, \dots, \mathbf{v_{ij}}) \\
        \mathbf{m_i} &= \text{BiLSTM}(\mathbf{m_{i1}}, \dots, \mathbf{m_{ik}}) \\
        \hat{t_i},\hat{v_i} &= \text{BiLSTM}(\mathbf{t_i}, \mathbf{v_i})  \\
        \hat{r_i},\hat{m_i} &= \text{BiLSTM}(\mathbf{r_i}, \mathbf{m_i}) 
\end{align*}
Figure \ref{fig:encoder} shows an example of the encoder.

\subsection{Decoder}

We represent the source code to be produced as a sequence of production rules ($a_t$ at step $t$), with a non-terminal $n_t$ on the left hand side and a combination of terminal and non-terminal symbols on the right hand side (see Figure \ref{fig:task}). The first non-terminal is \textit{MemberDeclaration}.  Subsequently, every non-terminal is expanded in a depth first left to right fashion, similar to \newcite{yin-neubig:2017:Long}. The probability of a source code snippet is decomposed as a product of the conditional probability of generating each step in the sequence of rules conditioned on the previously generated rules. Our decoder is an LSTM-based RNN that produces a context vector $c_t$ at each time step, which is used to compute a distribution over next actions. 
\begin{align} \label{eq:1}
  p(a_t|a_{<t}) &\propto exp(W^{n_t} c_t)
\end{align}
Here, $W^{n_t}$ is a $|n_t| \times H$ matrix, where $|n_t|$ is the total number of unique production rules that $n_t$ can be expanded to. The context vector $c_t$ is computed using the hidden state $s_t$ of an n-layer decoder LSTM cell and attention vectors over the NL and the context ($z_t$ and $e_t$), as described below. 

\paragraph{Decoder LSTM} The decoder uses an n-layer LSTM whose hidden state $s_t$ is computed based on the current non-terminal $n_t$ to be expanded, the previous production rule $a_{t-1}$, the parent production rule par$(n_t)$ that produced $n_t$, the previous decoder LSTM state $s_{t-1}$, and the decoder state of the LSTM cell that produced $n_t$, denoted as $s_{n_t}$. \begin{align}
  s_t &= \text{LSTM}(n_t, a_{t-1}, \text{par}(n_t), s_{t-1}, s_{n_t})
\end{align}
We use an embedding matrix $N$ to embed $n_t$ and matrix $A$ to embed $a_{t-1}$ and $\text{par}(n_t)$. If $a_{t-1}$ is a rule that generates a terminal node that represents an identifier or literal, it is represented using a special rule \textit{IdentifierOrLiteral} to collapse all these rules into a single previous rule. 

\paragraph{Two-step Attention} At time step $t$, the decoder first attends to every token in the NL representation, $h_i$, using the current decoder state, $s_t$, to compute a set of attention weights $\alpha_t$, which are used to combine $h_i$ into an NL context vector $z_t$. We use a general attention mechanism~\cite{luong-pham-manning:2015:EMNLP}, extended to perform multiple steps.
\begin{align*}
\alpha_{t, i} &= \frac{\text{exp}(s_t^\text{T} \mathbf{F} h_i)}{\sum_{i} \text{exp}(s_t^\text{T} \mathbf{F} h_i)} \\
z_t &=  \sum_{i} \alpha_{t, i} h_i
\end{align*}
The context vector $z_t$ is used to attend over every type (return type) and variable (method) name in the environment, to produce attention weights $\beta_t$ that are used to combine the entire context \mbox{$x=[t:v:r:m]$} into an environment context vector $e_t$.\footnote{``:'' denotes concatenation.}
\begin{align*}
\beta_{t, j} &= \frac{\text{exp}(z_t^\text{T} \mathbf{G} x_j)}{\sum_{j} \text{exp}(z_t^\text{T} \mathbf{G} x_j)} \\
e_t &=  \sum_{j} \beta_{t, j} x_j
\end{align*}
Finally, $c_t$ is computed using the decoder state and both context vectors $z_t$ and $e_t$:
\begin{align*}
  c_t = tanh(\hat{W}[s_t:z_t:e_t])
\end{align*}

\begin{figure}
    \includegraphics[width=\linewidth]{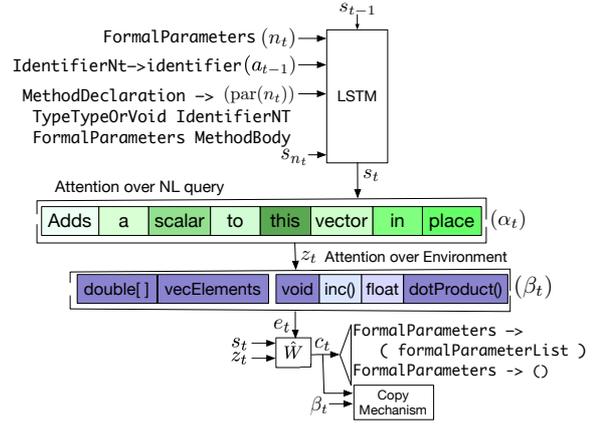}
    \caption{The hidden state $s_t$ of our decoder is a function of the previous hidden state, current non-terminal, previous production rule,  parent rule, and the parent hidden state. $s_t$ is used to attend on the NL and compress it into $z_t$, which is then used to attend over the environment variables and methods to generate $e_t$. The decoder uses all these context vectors to produce a distribution over valid right hand side values of the current non-terminal, and also learns to copy from the environment.}
    \label{fig:decoder}
\end{figure}

\paragraph{Supervised Copy Mechanism} Since the class environment at test time can belong to previously unseen new domains, our model needs to learn to copy variables and methods into the output. We use the copying technique of \newcite{gu-EtAl:2016:P16-1} to compute a copy probability at every time step $t$ using vector $b$ of dimensionality $H$.
\begin{align*}
   \text{copy}(t) = \sigma(b^T c_t)
\end{align*}
Since we only require named identifiers or user defined types to be copied, both of which are produced by production rules with $n_t$ as \textit{IdentifierNT}, we make use of this copy mechanism only in this case. Identifiers can be generated by directly generating derivation rules (see equation \ref{eq:1}), or by copying from the environment. The probability of copying an environment token $x_j$, is set to be the attention weights $\beta_{t,j}$ computed earlier, which attend exactly on the environment types and names which we wish to be able to copy. The copying process is supervised by preprocessing the production rules to recognize identifiers that can be copied from the environment, and both the generation and the copy probabilities are weighted by $1 - \text{copy}(t)$ and $\text{copy}(t)$ respectively. The LSTM decoder with attention mechanism is illustrated in Figure \ref{fig:decoder}.

\subsection{Baseline Models}

\paragraph{Retrieval} We evaluate a retrieval baseline, where the output source code for a test example is the training example source code whose NL is closest in terms of cosine similarity to the test NL using a tf-idf representation. We then replace all occurrences of environment variables and methods in the chosen training source code with similarly typed variables and methods from the environment of the test example, and we break ties randomly.

\paragraph{Seq2seq} We evaluate a Seq2Seq baseline by representing the NL and context as a sequence formed by the concatenation of the NL, the variables and the methods with separators between them. The variables (methods) are represented with the type (return type) followed by the name, with a different separator between them. The encoder is an n-layer LSTM which initializes an LSTM-based decoder using its final hidden states. The decoder uses an attention mechanism \cite{luong-pham-manning:2015:EMNLP} over the encoder states to produce a conditional distribution over the next source code token (not production rule) given all the previous tokens. We replace UNK tokens in the output with source tokens having the most attention weight.
We also attempted to evaluate the Seq2Tree model of \newcite{dong2016} but the redundancy in the model resulted in extremely long output sequences which did not scale. Experiments on a smaller dataset gave comparable results to Seq2seq. 

\paragraph{Seq2prod} This baseline corresponds to the action sequence model by \newcite{yin-neubig:2017:Long}, with a BiLSTM  over a sequence representation of the NL and context (same as Seq2seq), and a decoder that learns to generate a derivation of the AST of the underlying source code, similar to our model. The decoder uses the same attention mechanism as the Seq2seq, however, it uses supervised copying from the entire input sequence to handle  unknown words encountered during testing. 


\section{CONCODE}

\begin{table}[t]
\centering
\begin{tabular}{lr} 
\toprule
  & \textbf{Count} \\
\midrule
Train & 100,000 \\
Valid/Test & 2,000/2,000 \\
Average NL Length  & 13.73 \\
Average Code Characters  & 119 \\
Average Code Tokens  & 26.3 \\
Average \# Environment Variables & 4.89 \\
Average \# Environment Methods & 10.95 \\
\midrule
Average AST Nodes & 79.6 \\
\# Node Types & 153 \\
\# Production Rules & 342* \\
\midrule
\% Getters & 16.74\% \\
\% Setters & 3.39\% \\
\% using Class Variables & 68\%\\
\% using Class Methods  & 16.2\%\\
\% Unknown Types  & 7.65\%\\
\bottomrule
\end{tabular}
\caption{Statistics of our dataset of (NL, context and code) tuples collected from Github. *Does not include rules that generate identifiers.}
\label{dataset}
\end{table}

We built a new dataset (CONCODE) from public Java projects on Github that contains environment information together with NL (Javadoc-style method comments) and code tuples. We gather Java files from $\sim$33,000 repositories, which are then split into train, development, and test sets based on repository, rather than purely randomly. Dividing based on the repository keeps the domains in the test set separate from the training set, therefore providing near zero-shot conditions that should truly test the ability of models to generalize to associate unseen NL tokens with previously unseen environment variables and methods. We also remove all examples from the development and test sets where the NL is exactly present in the training set. We further eliminate all Java classes that inherit from parent classes, since the resulting code may use variables and methods inherited from parent classes that reside in separate source files. While this is an important and interesting feature, we leave it for future work. 

Every method that contains a Javadoc comment is treated as a training example. The Javadoc comment forms the NL, the method body is the target code to be generated, and all member variables, as well as other member method signatures are extracted and stored as part of the context. The Javadoc is preprocessed by stripping special symbols such as \texttt{@link}, {\tt @code}, {\tt @inheritDoc} and special fields such as \texttt{@params}. Methods that do not parse are eliminated and the rest are pre-processed by renaming locally defined variables canonically, beginning at \textit{loc0} and similarly for arguments, starting with \textit{arg0}.  We also replace all method names with the word \textit{function} since it doesn't affect the semantics of the resulting program. Generating informative method names has been studied by \newcite{allamanis2015suggesting}. 
We replace all string literals in the code with constants as they are often debug messages. Finally, we use an ANTLR java grammar\footnote{\tt https://github.com/antlr/grammars-v4} that is post-processed by adding additional non-terminals and rules to eliminate wildcard symbols in the grammar, in order to convert the source code into a sequence of production rules.
The resulting dataset contains 100,000 examples for training, and 2000 examples each for development and testing, respectively. Table \ref{dataset} summarizes the various statistics. We observe that on average, an environment contains $\sim$5 variables and $\sim$11 methods. Around $68\%$ of the target code uses class member variables, and $16\%$ of them use member methods, from the environment. Based on a frequency cutoff of 2 on the training set, we find that $7\%$ of the types in the development set code are unknown, hence they need to be copied from the environment. Since CONCODE is extracted from a diverse set of online repositories, there is a high variety of code templates in the dataset compared to existing datasets. For example, a random baseline on the Hearthstone card game dataset \cite{ling2016} gives a BLEU score of 40.3, but only a score of 10.2 on CONCODE. We plan to release all code and data from this work.\footnote{\tt https://github.com/sriniiyer/concode}

\section{Experimental Setup} 

We restrict all models to examples where the length of the combination of the NL and the context is at most $200$ tokens and the length of the output source code is at most $150$ tokens. Source NL tokens are lower-cased, camel-cased identifiers are split and lower-cased, and are used together with the original version. The vocabulary for identifiers uses a frequency threshold of $7$, resulting in a vocabulary of $32,600$ tokens. The types vocabulary uses a threshold of $2$ resulting in $22,324$ types. We include all $153$ non-terminals and $342$ previous rules. We use a threshold of $2$ for output production rules to filter out the long tail of rules creating identifiers and literals, resulting in $18,135$ output rules. Remaining values are replaced with the UNK symbol.

\paragraph{Hyperparameters} We use an embedding size $H$ of 1024 for identifiers and types. All LSTM cells use 2-layers and a hidden dimensionality of 1024 (512 on each direction for BiLSTMs). We use an embedding size of $512$ for encoding non-terminals and rules in the decoder. We use dropout with $p=0.5$ in between LSTM layers and at the output of the decoder over $c_t$. We train our model for $30$ epochs using mini-batch gradient descent with a batch size of $20$, and we use Adam \cite{kingma2014adam} with an initial learning rate of $0.001$ for optimization. We decay our learning rate by $80\%$ based on performance on the development set after every epoch.  

\paragraph{Inference and Metrics} Inference is done by first encoding the NL and context of the test example. We maintain a stack of symbols starting with the non-terminal, \textit{MemberDeclaration}, and at each step, a non-terminal (terminals are added to the output) is popped off the stack to run a decoding step to generate the next set of symbols to push onto the stack. The set of terminals generated along the way forms the output source code. We use beam search and maintain a ranked list of partial derivations (or code tokens for Seq2seq) at every step to explore alternate high-probability derivations. We use a beam size of $3$ for all neural models. We copy over source tokens whenever preferred by the model output. We restrict the output to $150$ tokens or $500$ production rules. 

To evaluate the quality of the output, we use Exact match accuracy between the reference and generated code. As a measure of partial credit,
we also compute the BLEU score \cite{papineni2002bleu}, following recent work on code generation~\cite{ling2016, yin-neubig:2017:Long}. BLEU is an n-gram precision-based metric that will be higher when more subparts of the predicted code match the provided reference. 

\begin{table}
\centering
\begin{tabular}{lll} 
\toprule
\textbf{Model} &  Exact & BLEU\\
\midrule
Retrieval & 2.25 (1.65) & 20.27 (18.15) \\
Seq2Seq & 3.2 (2.9) & 23.51 (21.0) \\
Seq2Prod & 6.65 (5.55) & 21.29 (20.55)  \\ 
Ours & 8.6 (7.05) & 22.11 (21.28)  \\
\bottomrule
\end{tabular}
\caption{Exact match accuracy and BLEU score (for partial credit) on the test (development) set, comprising 2000 examples from previously unseen repositories.}
\label{fig:results}
\end{table}

\begin{table}
\centering
\begin{tabular}{lll} 
\toprule
\textbf{Model} & Exact & BLEU   \\
\midrule
Ours  &  7.05 & 21.28  \\
-Variables  & 1.6 & 20.78    \\
-Methods & 6.2 & 21.74  \\
-Two step attention & 5.75 & 17.2  \\
-Camel-case encoding & 5.7 & 21.83  \\
\bottomrule
\end{tabular}
\caption{Ablation of model features on the development set.}
\label{ablations}
\end{table}

\begin{figure*}
    \includegraphics[width=\linewidth]{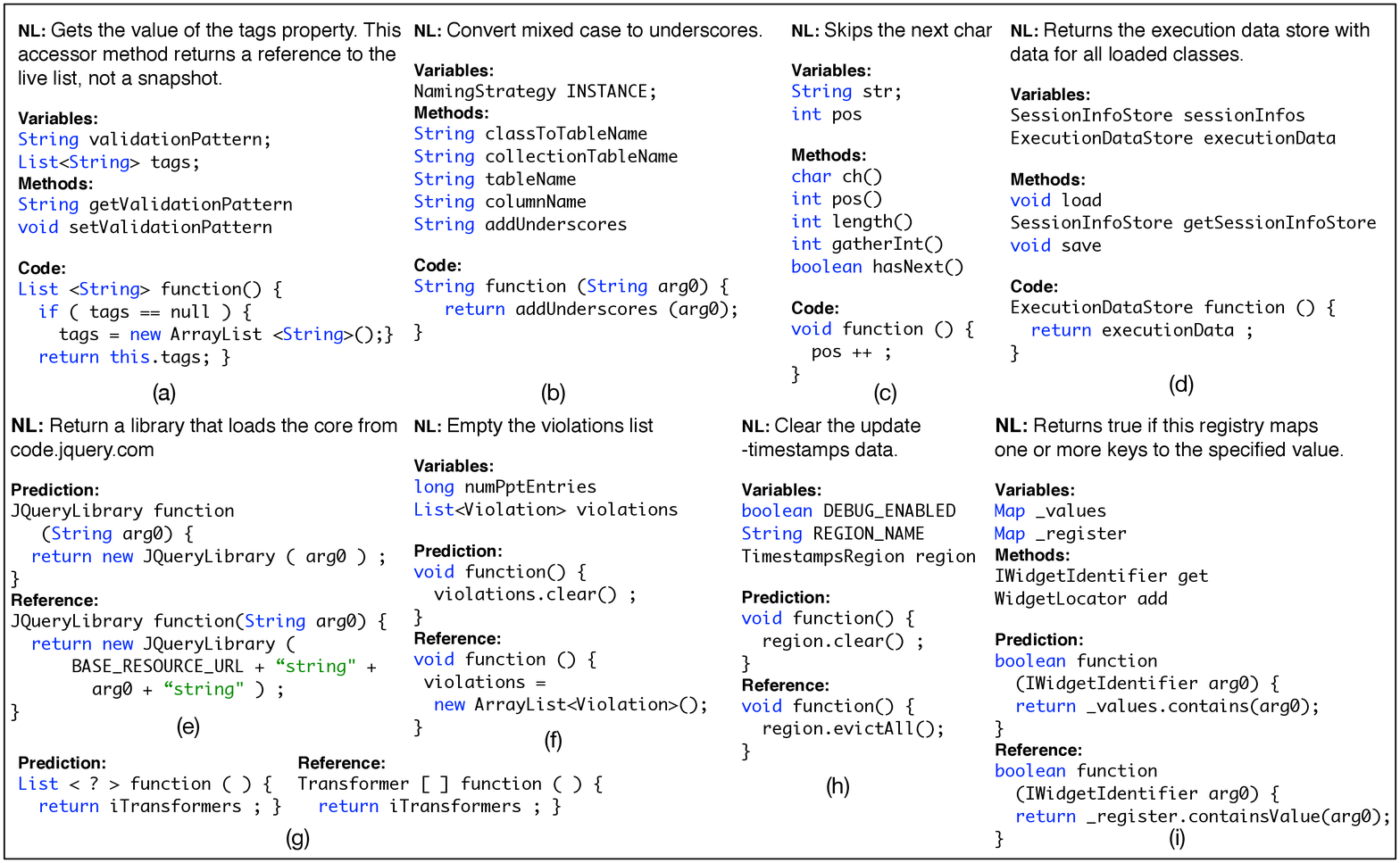}
    \caption{Analysis of our model output on development set examples. Some environment variables and methods are omitted for space. (a)-(d) represent cases where the model exactly produced the reference output. (e)-(g) are cases where the model output is very reasonable for a practical setting. In (f), the model produces a better solution than the reference. In (h), the context lacks information to produce the reference code. The model chooses the wrong element in (i) and could be improved by better encoder representations.}
    \label{fig:examples}
\end{figure*}

\section{Results} We present results for the context based code generation task on the test and dev sets in Table \ref{fig:results}. Our model outperforms all baselines and achieves a gain of 1.95\% exact match accuracy and 0.82 BLEU points, with respect to the next best models. The combination of independently encoding sub-word units and applying a two-step attention mechanism helps the model learn to better associate the correct variables/methods from the context and the language in the NL. Figure \ref{fig:examples} (a) shows an example output of our model, which produces code structure intermixed with member variables (tags). In (b) our model learns to call method \texttt{addUnderscores} (an UNK in the vocabulary) with its correct return type ({\tt String}). Similarly, in (d) our model also successfully learns to use a previously unseen type (\texttt{ExecutionDataStore}) when making use of the corresponding variable. (c) is an example of where the NL does not directly refer to the variable to be used. The mismatch between dev and test results is because we ensure that the dev and test examples come from non-overlapping Github repositories, resulting in different distributions.

Using a constrained decoder that generates syntax tree rules instead of tokens leads to significant gains in terms of exact match score (6.65 for Seq2prod vs 3.2 for Seq2seq), and shows that this is an important component of code generation systems. Seq2prod, however, fails on examples (b)-(d), since it is harder to learn to match the NL tokens with environment elements. Finally, all neural models outperform the retrieval baseline with member substitution. 

To understand which components of the data and the model contribute most to the output, we perform an ablation study on the development set (Table \ref{ablations}). Removing the variables leads to a significant hit in exact match accuracy since 68\% of examples (Table \ref{dataset}) refers to class variables; a similar but smaller reduction is incurred by removing methods. The presence of these components makes this task more challenging and also more aligned with programming scenarios in practice. Removing the two-step attention mechanism leads to a 1.3\% drop in accuracy since the attention on the NL is unable to interact with the attention on the environment variables/methods. Removing camel-case encoding leads to a small drop mainly because many variable (method) names are single words.

\section{Error Analysis}

Subfigures \ref{fig:examples}(e)-(i) show cases where our model output did not exactly match the reference. In (e)-(g), the model output is semantically equivalent to the reference and is a very reasonable prediction in a practical setting. 
For example, in (e) the only difference between the prediction and the reference is the string concatenations to the url. Interestingly, in example (f) the prediction is a cleaner way to achieve the same effect as the reference, and this is a great example of the application of these models for suggesting standard and efficient code. Unfortunately, our model is penalized by the exact match metric here. Similarly, in (g), the model uses a generic list (\texttt{List<?>}) in place of the specific type (\texttt{Transformer[]}). Example (h) demonstrates a case where the model is unaware of methods that can be called on class members (specifically that \texttt{evictAll} is a member of the \texttt{TimestampsRegion} class), and requires augmenting the environment with additional member type documentation, which we believe will be an important area for future work. Example (i) calls for richer encoder representations, since our model incorrectly uses the \texttt{\_values} variable instead of \texttt{\_register}, as it is unable to associate the word ``registry'' with the right elements. 

We further perform a qualitative analysis of 100 predictions on our development set (Table \ref{tab:qualitative}) and find that there is significant room for improvement with 71\% of the predictions differing significantly from their references. 16\% of the predictions are very close to their references with the difference being 1-2 tokens, while 11\% are exactly correct. 2\% of the predictions were semantically equivalent but not exactly equal to their references.

\begin{table}
\centering
\begin{tabular}{lll} 
\toprule
\textbf{Category} & \textbf{Fraction} \\
\midrule
Totally Wrong  &  62\%  \\
Marginally Correct  & 9\% \\
Mostly Correct & 16\%  \\
Exact Match & 11\%  \\
Semantically Equivalent & 2\%  \\
\bottomrule
\end{tabular}
\caption{Qualitative distribution of errors on the development set. }
\label{tab:qualitative}
\end{table}

\section{Related Work}

There is significant existing research on mapping NL directly to executable programs in the form of logical forms \cite{zettlemoyer05}, $\lambda$-DCS \cite{liang2011learning}, regular expressions \cite{kushman-barzilay:2013:NAACL-HLT,locascio-EtAl:2016:EMNLP2016}, database queries \cite{iyer-EtAl:2017:Long,zhong2017seq2sql} and general purpose programs \cite{balog2016deepcoder,allamanis2015bimodal}. 
\newcite{ling2016} generate Java and Python source code from NL for card games, conditioned on categorical card attributes. \newcite{manshadi2013integrating} generate code based on input/output examples for applications such as database querying. 
\newcite{Gu:2016:DAL:2950290.2950334} use neural models to map NL queries to a sequence of API calls, and 
\newcite{neelakantan2015neural} augment neural models with a small set of basic arithmetic and logic operations to generate more  meaningful programs. 
In this work, we introduce a new task of generating programs from NL based on the environment in which the generated code resides, following the frequently occurring pattern in large repositories where the code depends on the types and availability of variables and methods in the environment.

Neural encoder-decoder models have proved effective in mapping NL to logical forms and also for directly producing general purpose programs. 
\newcite{ling2016} use a sequence-to-sequence model with attention and a copy mechanism to generate source code. Instead of directly generating a sequence of code tokens, recent methods focus on constrained decoding mechanisms to generate syntactically correct output using a decoder that is either grammar-aware or has a dynamically-determined modular structure paralleling the structure of the abstract syntax tree (AST) of the code \cite{dong2016,rabinovich-stern-klein:2017:Long,krishnamurthy-dasigi-gardner:2017:EMNLP2017,yin-neubig:2017:Long}.
Our model also uses a grammar-aware decoder similar to \newcite{yin-neubig:2017:Long} to generate syntactically valid parse trees, augmented with a two-step attention mechanism \cite{chen2016thorough}, 
followed by a supervised copying mechanism \cite{gu-EtAl:2016:P16-1} over the class environment. 

Recent models for mapping NL to code have been evaluated on datasets containing highly templated code for card games (Hearthstone \& MTG; \citeauthor{ling2016}, \citeyear{ling2016}), or manually labeled per-line comments (DJANGO; \citeauthor{oda2015learning}, \citeyear{oda2015learning}). These datasets contain $\sim$20,000 programs with short textual descriptions possibly paired with categorical data, whose values need to be copied onto the resulting code from a single domain. In this work, we collect a new dataset of over 100,000 NL and code pairs, together with the corresponding class environment. 
Each environment and NL  describe a specific domain and the dataset comprises thousands of different domains, that poses additional challenges. Having an order of magnitude more data than existing datasets makes training deep neural models very effective, as we saw in the experimental evaluation. While massive amounts of Github code have been used before for creating datasets on source code only \cite{allamanis2013mining,allamanis2014mining,allamanis2016convolutional},
we instead extract from Github a dataset of NL and code with an emphasis on context, in order to learn to map NL to code within a class.  


\section{Conclusion}

In this paper, we introduce new data and methods for learning to generate source code from language within the context of a real-world code base. 
To train models for this task, we collect and release CONCODE, a large new dataset of NL, code, and context tuples from online repositories, featuring code from a variety of domains. We also introduced a new  encoder decoder model with a specialized context encoder
which outperforms strong neural baselines by 1.95\% exact match accuracy. Finally, analysis suggests that even richer models of programmatic context could further improve these results.

\section*{Acknowledgements}

The research was supported in part by DARPA (FA8750-16-2-0032), the ARO (ARO-W911NF-16-1-0121), the NSF (IIS-1252835, IIS-1562364, IIS1546083, IIS-1651489, OAC-1739419), the DOE (DE-SC0016260), the Intel-NSF CAPA center, and gifts from Adobe, Amazon, Google, and Huawei. The authors thank the anonymous reviewers for their helpful comments.

\bibliography{emnlp2018}
\bibliographystyle{acl_natbib_nourl}

\end{document}